\documentclass{article} % For LaTeX2e
\usepackage{iclr2026_conference,times}
\usepackage{microtype}

\usepackage{amsmath,amsfonts,bm}

\def\eqref#1{equation~\ref{#1}}
\def\1{\bm{1}}

\DeclareMathAlphabet{\mathsfit}{\encodingdefault}{\sfdefault}{m}{sl}
\SetMathAlphabet{\mathsfit}{bold}{\encodingdefault}{\sfdefault}{bx}{n}

\usepackage{url}
\usepackage{algorithm, algorithmic}

\usepackage{multirow}
\usepackage{booktabs}
\usepackage{graphicx}

\usepackage{listings}
\title{Beyond the Answer: Decoding the Behavior of LLMs as Scientific Reasoners}

\author{% need this comment here to comment out the newline hehe :)
\textbf{Rohan Pandey}\thanks{Equal contribution} \quad \textbf{Eric Ye}\footnotemark[1] \\
University of Washington \\
\texttt{\{rpande, ericy4\}@uw.edu} \\
\And
\textbf{Michael Li} \footnotemark[1] \\
Carnegie Mellon University \\
\texttt{ml7@andrew.cmu.edu}
}

\iclrfinalcopy % Uncomment for camera-ready version, but NOT for submission.
\begin{document}

\maketitle

\begin{abstract}
As Large Language Models (LLMs) achieve increasingly sophisticated performance on complex reasoning tasks, current architectures serve as critical proxies for the internal heuristics of frontier models. Characterizing emergent reasoning is vital for long-term interpretability and safety. Furthermore, understanding how prompting modulates these processes is essential, as natural language will likely be the primary interface for interacting with AGI systems. In this work, we use a custom variant of Genetic Pareto (GEPA) to systematically optimize prompts for scientific reasoning tasks, and analyze how prompting can affect reasoning behavior. We investigate the structural patterns and logical heuristics inherent in GEPA-optimized prompts, and evaluate their transferability and brittleness. Our findings reveal that gains in scientific reasoning often correspond to model-specific heuristics that fail to generalize across systems, which we call ``local'' logic. By framing prompt optimization as a tool for model interpretability, we argue that mapping these preferred reasoning structures for LLMs is an important prerequisite for effectively collaborating with superhuman intelligence.
\end{abstract}

\section{Introduction}

As the capabilities of Large Language Models (LLMs) increasingly push performance frontiers \citep{bubeck2023sparks}, research focus will foreseeably shift from benchmark tracking toward understanding how to best collaborate with LLMs. We identify reasoning as the most critical capability of LLMs in a post-AGI landscape, as it provides the verifiable logical scaffolding necessary for autonomous systems to navigate novel, high-stakes scenarios \citep{huang2023reasoninglargelanguagemodels}. A rigorous understanding of these mechanisms enables more effective human-AGI collaboration in a wide range of scenarios, ensuring these systems act reliably when tackling complex challenges. Current LLMs provide a valuable window into the reasoning paradigms that may define a post-AGI world.

We see scientific reasoning as a robust testbed for probing these internal paradigms. Scientific reasoning provides essential foundational logic and structured frameworks, which can generalize to a variety of high impact downstream tasks, including but not limited to scientific discovery, spatial navigation, and engineering. In this work, we examine such reasoning in two domains: (1) GPQA, which tests graduate-level scientific reasoning \citep{rein2023gpqa}, and (2) a formally verified algebra dataset implemented in Lean \citep{yang2023leandojo, zheng2022minif2f}. These benchmarks allow for the observation of complex logic under controlled and interpretable conditions.

Since natural language will likely be the primary interface for interacting with AGI systems, understanding how prompting affects reasoning capabilities is crucial for reliable collaboration. In this work, we first use Genetic Pareto (GEPA) to optimize prompts and discover which specific instructions result in the best performance \citep{agrawal2025gepa, khattab2023dspy}. We then analyze these results to identify the patterns that elicit higher-level reasoning in these models.

We find that high-performing prompts often rely on model-specific heuristics that fail to generalize across models \citep{mirzadeh2025gsmsymbolic}. Mapping these machine-preferred structures is vital for overseeing future general-purpose systems \citep{berglund2024reversalcurse}. This work establishes a foundation for decoding LLM reasoning to ensure safety frameworks are prepared for a post-AGI society.

\section{Background}

\paragraph{LLMs for Math and Science}

LLMs are rapidly moving beyond simple linguistic pattern matching, developing complex multi-step reasoning skills that are necessary for mathematical and scientific general intelligence. Current methodologies often rely on specialized prompting paradigms; for instance, Chain-of-Thought \citep{wei2022chain} encourages explicit symbolic derivation, while Program-of-Thought \citep{chen2022program} offloads complex scientific computing to external Python interpreters. More recently, reasoning-centric models such as OpenAI’s o1 and DeepSeek’s R1 have utilized large-scale reinforcement learning to internalize these logical trajectories \citep{deepseekr12025}. However, breaking records on benchmarks is only a partial milestone. As these systems approach AGI-level performance, research must shift from merely tracking performance to understanding how to best collaborate with these models. In this work, we argue that identifying the structural biases and implicit reasoning strategies within these models is a vital prerequisite for the effective oversight and safe deployment of AGI systems in high-stakes and unsupervised scenarios.

\paragraph{Automated Prompt Engineering}

Optimization of model performance has evolved from the manual engineering of prompts to a systematic algorithmic search. Early works in automated prompt engineering have demonstrated that LLMs are often the best optimizers of their own instructions \citep{zhou2022large, yang2023large}. Genetic Pareto (GEPA) utilizes an evolutionary approach to iteratively optimize high performing prompts \citep{agrawal2025gepa}. While previous works on automated engineering focused primarily on maximizing accuracy, we reposition these algorithms as tools for understanding. By allowing GEPA to explore the vast search space of possible instructions, we treat the resulting optimized prompts and the trajectory of prompt evolution as valuable windows into the latent preferences of the model. We can then identify specific heuristics that can be reverse engineered and transferred back to improve human-authored prompting methodologies.

\paragraph{Knowledge Transferability}

A fundamental question in the path toward AGI is whether machine intelligence is a singular phenomenon or a collection of fragmented epistemologies \citep{quattrociocchi2025episfaultlines}. Previous research into model distillation and generalization has shown that while knowledge can be transferred between architectures, internal reasoning protocols often remain model specific \citep{hinton2015distilling}. This brittleness poses challenges for the post-AGI era. For AI systems to truly be of use in collaborative scientific discovery, their logic must be interoperable. If a reasoning strategy optimized for one model fails on another, it suggests a closed epistemology, which represents a detached form of intelligence that lacks a universal logical foundation \citep{pal2026explanationsgeneralize}.

\section{Methodology}

We assess LLM reasoning in two domains: (1) formal mathematical theorem proving via Lean from the MiniF2F dataset \citep{zheng2022minif2f} and (2) scientific reasoning via the GPQA Diamond benchmark \citep{rein2023gpqa}. We refer to these benchmarks as ``Algebra'' and ``GPQA'' respectively. These domains assess scientific reasoning in complementary ways: Lean requires rigid verifiable logic and is open ended, while GPQA evaluates high level conceptual reasoning and is multiple choice. We apply a custom variant of GEPA which has been simplified and adapted for Lean theorem proving and GPQA. Implementation details are specified in Algorithm \ref{alg:pipeline}.

\begin{algorithm}[htbp]
\caption{Custom Variant of Genetic Pareto (GEPA)}
\label{alg:pipeline}
\begin{algorithmic}[1]
\STATE \textbf{Input:} Seed prompt $P_0$, Lean theorems $L$, GPQA questions $G$, iterations $T$, samples $(n, m)$.
\STATE \textbf{Initialize:} $Population \gets \{P_0\}$, $Pareto \gets \emptyset$.
\FOR{$t = 1$ \TO $T$}
    \STATE $P \gets \text{Sample}(Pareto \neq \emptyset \ ? \ Pareto : Population)$
    \STATE $S_L \gets \text{Sample}(L, n)$, $S_G \gets \text{Sample}(G, m)$
    \STATE \textbf{Evaluate:} $\mathbf{v}[i] \gets \begin{cases} \text{Lean\_Verify}(\text{LLM}(P, c_i)) & c_i \in S_L \\ \text{Check\_Answer}(\text{LLM}(P, c_i)) & c_i \in S_G \end{cases}$ for $c_i \in S_L \cup S_G$
    \STATE \textbf{Update Pareto:} Add $P$ if non-dominated by existing prompts; remove dominated.
    \STATE $Errors \gets \{c_i : \mathbf{v}[i] = 0\}$
    \STATE $Critique \gets \text{LLM\_Critic}(P, \text{Logs}(Errors))$
    \STATE $P' \gets \text{LLM\_Evolve}(P, Critique)$
    \STATE $Population \gets Pareto \cup \{P'\}$ \COMMENT{Prune to Pareto + new child}
\ENDFOR
\RETURN $Pareto$
\end{algorithmic}
\end{algorithm}

We conduct the entirety of GEPA optimization using DeepSeek-V3.2, which currently has state-of-the-art performance on reasoning benchmarks. To gain insight into the transferability of optimized prompts across different models, we also test the same prompts on ChatGPT-5.4-mini, GLM 5, and Claude Sonnet 4.6, all of which were released within the past few months and have competitive performance. By analyzing the prompt optimization process and comparing prompt performance across models, we hope to gain insight into how prompting strategy affects reasoning performance.

For every combination of model and benchmark, we run evaluations across four prompts: \textbf{(1) Hand-Crafted Simple:} A hand-crafted simple prompt, intended to emulate the prompting ability of average technical users, which serves as a baseline; \textbf{(2) Hand-Crafted CoT:} A hand-crafted Chain-of-Thought prompt, intended to emulate best practice prompting strategies to current knowledge, which serves as a baseline; \textbf{(3) GEPA Optimized Baseline:} The initial prompt from the GEPA optimization process; and \textbf{(4) GEPA Optimized Final:} The final prompt from the GEPA optimization process. Prompts are further discussed in the Appendix, and examples are provided.

\section{Results}

\subsection{Benchmark Performance}

\begin{table}[htbp]
  \caption{Performance comparison of models on benchmark datasets.}
  \label{tab:results}
  \centering
  \begin{tabular}{llll}
    \toprule
    \textbf{Model} & \textbf{Method} & \textbf{Algebra} & \textbf{GPQA} \\
    \midrule
    \multirow{4}{*}{DeepSeek-V3.2} 
            & Hand-Crafted Simple & 86.11\% & 91.67\% \\
            & Hand-Crafted CoT & 97.22\% & 91.67\% \\
            & GEPA Optimized Baseline & 91.67\% & 88.89\% \\
            & GEPA Optimized Final & \textbf{100.00\%} & \textbf{94.44\%} \\
    \midrule
    \multirow{4}{*}{GPT-5.4-mini} 
            & Hand-Crafted Simple & 50.00\% & \textbf{91.67\%} \\
            & Hand-Crafted CoT & 47.22\% & \textbf{91.67\%} \\
            & GEPA Optimized Baseline & 50.00\% & 88.89\% \\
            & GEPA Optimized Final & \textbf{61.11\%} & \textbf{91.67\%} \\
    \midrule
    \multirow{4}{*}{GLM 5} 
            & Hand-Crafted Simple & 91.67\% & \textbf{91.67\%} \\
            & Hand-Crafted CoT & \textbf{97.22\%} & 86.11\% \\
            & GEPA Optimized Baseline & 91.67\% & 88.89\% \\
            & GEPA Optimized Final & 94.44\% & \textbf{91.67\%} \\
    \midrule
    \multirow{4}{*}{Claude Sonnet 4.6} 
            & Hand-Crafted Simple & 30.56\% & 77.78\% \\
            & Hand-Crafted CoT & \textbf{52.78\%} & \textbf{83.33\%} \\
            & GEPA Optimized Baseline & 50.00\% & 80.56\% \\
            & GEPA Optimized Final & 50.00\% & 80.56\% \\
    \bottomrule
  \end{tabular}
\end{table}

Prompt performance on benchmarks is displayed in Table \ref{tab:results}. Our first major observation is that the GEPA Optimized Final prompt achieves its most significant gains on DeepSeek-V3.2, reaching 100.00\% on Algebra and 94.44\% on GPQA. This confirms that the GEPA optimization process is highly effective when evaluated on the same model used during optimization. Our second major observation is that the superiority of the GEPA-optimized prompts does not reliably transfer to other models. While some models show marginal benefits -- for example, GPT-5.4-mini sees an improvement in Algebra from 50.00\% to 61.11\% -- the optimized prompt is rarely the undisputed best performer elsewhere. Notably, for GLM 5 (Algebra) and Claude Sonnet 4.6 (both benchmarks), the Hand-Crafted CoT prompts actually outperform the GEPA Optimized Final prompt. 

These results suggest that the optimization process is highly model-specific, and suggests a trade-off between prompt universality and performance. Since DeepSeek was used as the optimizer, the evolved prompts likely capture patterns specific to DeepSeek's architecture that do not resonate with the internal logic of other models. This highlights a significant lack of interoperability and reinforces the brittleness of automated prompt engineering across diverse foundation models.

\begin{figure}[htbp]
  \centering
  \includegraphics[width=0.8\linewidth]{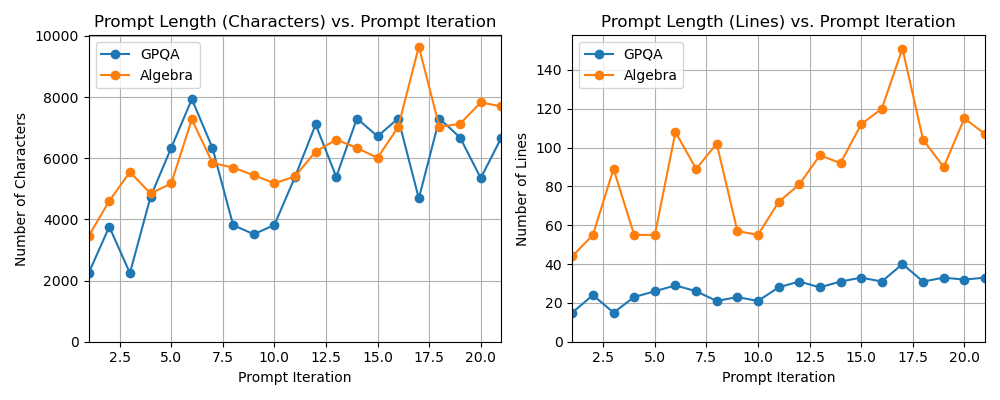}
  \caption{The length of GEPA proposed prompts increases over the course of optimization, with the final prompt often being about twice as long in characters as the initial prompt. This shows that detailed prompting is likely required to unlock better reasoning capabilities in LLMs.}
  \label{fig:prompt_lengths}
\end{figure}

\subsection{Prompt Evolution Patterns}

By manually analyzing the evolution of prompts over the course of GEPA optimization, we primarily find that prompts evolve from telling the model \emph{what to do} to coaching it on \emph{how to do it}, similar to a human expert. In the process, significant domain knowledge and best practices are often added to the context. For example, in the GEPA optimization process for the Algebra benchmark, prompts eventually referenced domain specific strategies such as using Eisenstein's Criterion for minimal polynomials. Similarly, later GPQA prompts mentioned specific strategies such as quantum field theory loop counting. Later Algebra prompts also explicitly warn against common pitfalls, especially with regard to Lean formatting. Interestingly, a robust protocol for ``Handling False Statements'' is also introduced to prevent hallucinating proofs. The Appendix contains prompt examples.

This observation is also supported by analyzing the prompt lengths over the course of optimization, as shown in Figure \ref{fig:prompt_lengths}. The final prompt turns out to be around twice as long as the initial prompt. This demonstrates that LLMs often rely on detailed prompting to fully elicit their reasoning capabilities.

By analyzing the embedding space for prompts over the optimization process, we also find that the embeddings of proposed prompts tends to drift in a consistent direction over the course of optimization. This suggests that for the same task, some regions of prompting space may offer more promising performance. A plot of this behavior is in the Appendix.

\section{Discussion}

\paragraph{Conclusion}

In this work, we demonstrated that using clever prompt optimization techniques can unlock reasoning potential in LLMs. We find that longer prompts, as well as prompts that guide the model along the \emph{how} of the target task, tend to elicit superior reasoning capabilities in LLMs. Such prompts can be optimized autonomously, and can beat hand-crafted baselines, even when we hand-craft prompts using state-of-the-art best practices. Our findings suggest that if current trends continue, post-AGI reasoning may not manifest as a universal logic, but rather as a collection of task-specific and model-specific heuristics that require precise coaching to elicit.

\paragraph{Limitations}

Our work is still relatively preliminary, and is limited to two benchmarks and four models. We would need a broader experimental scope to more confidently confirm the validity of our hypotheses. Furthermore, the stochastic nature of our LLM-driven prompt optimization loop could lead to inconsistency across multiple runs.

\paragraph{Future Work}

Our preliminary results suggest that prompt optimization remains dangerously architecture dependent. To prepare for a post-AGI landscape, future research should work on identifying ``reasoning primitives'', or logical structures invariant across models. If AGI develops logic that humans cannot understand, we need to build automated tools to keep these systems interpretable. Without this, we risk a future where our most capable tools are also our least predictable.

\section{Statements}

\subsection{Ethics Statement}

While the discovery of effective reasoning paths could potentially be repurposed for harmful tasks, our analysis of the logic inherent in these models primarily as a vital safeguard. If we better understand how LLMs reason, we can better utilize LLMs for reasoning related tasks. Ultimately, identifying how LLMs approach reasoning tasks is essential for the development of robust, aligned, and transparent post-AGI systems.

We also agree to NOT reveal examples from our evaluation datasets in plain text or images online, to reduce the risk of leakage into foundation model training corpora.

\subsection{Reproducibility Statement}

Our results are reproducible to the extent of nondeterminism in the outputs of LLMs used.

% Bibliography
\bibliography{iclr2026_conference}
\bibliographystyle{iclr2026_conference}

\appendix
\section{Appendix}

\subsection{Additional Results}

\begin{figure}[H]
  \centering
  \includegraphics[width=1.0\linewidth]{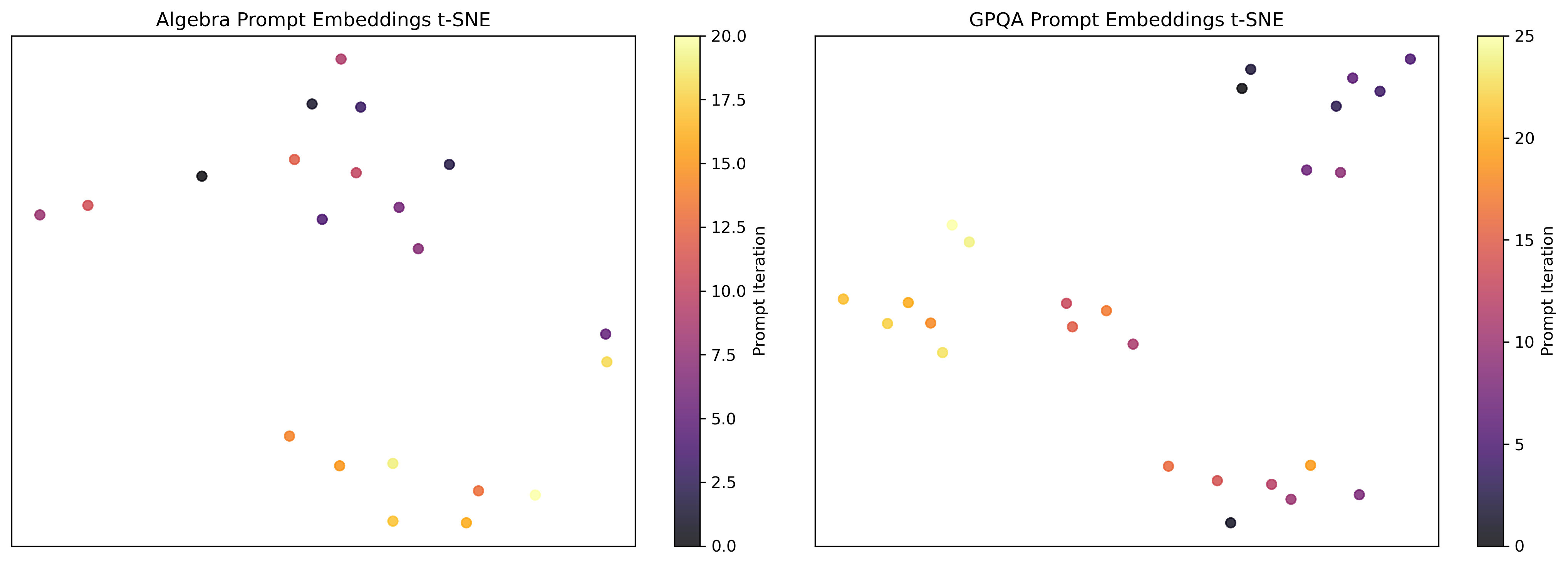}
  \caption{The embeddings of GEPA proposed prompts tend to drift over the course of optimization. For both algebra and GPQA, there seems to be a significant jump in embedding space at around iteration 12. This suggests that for the same task, some regions of prompting space may offer more promising performance.}
  \label{fig:prompt_embeddings}
\end{figure}

\subsection{Prompts}

The 4 prompts we tested for the Algebra benchmark are listed below. For sake of brevity, we do not include the 4 GPQA prompts in this section, but they are similar in nature.

\begin{lstlisting}[caption={Hand-Crafted Simple Prompt}, label={lst:prompt1}]
You are a mathematician and Lean 4 practitioner.
Your goal is to solve formal IMO-style problems and produce correct Lean 4 proofs.

Approach and tricks:
- Try to reduce the goal using simp/linarith/nlinarith or rewriting.
- Look for standard lemmas in Mathlib (algebra, number theory, inequalities, parity).
- Prefer structured proofs: have, refine, apply, and use `by` blocks.

Output format:
- Return the response in a single fenced code block with ```lean ... ``` containing valid Lean 4 code.
- Do not include any text outside the code block.
\end{lstlisting}
\begin{lstlisting}[caption={Hand-Crafted CoT Prompt}, label={lst:prompt2}]
You are an expert mathematician and Lean 4 developer.
Your goal is to solve formal IMO-style problems and produce correct Lean 4 proofs.

### Output Format
- Return the response in a single fenced code block with ````lean ... ```` containing valid Lean 4 code.
- Do not include any text outside the code block.
- The code should be complete, including necessary `import` and `open` statements.

### Imports
- Do not use a single monolithic `import Mathlib`.
- Instead, import the specific files you need, e.g., `import Mathlib.Data.Real.Basic`, `import Mathlib.Algebra.Order.Field.Basic`, `import Mathlib.Tactic.Linarith`.

### Chain-of-Thought (internal)
- Think step-by-step and consider small test cases for `ℕ`/`ℤ` statements.
- Do not reveal chain-of-thought. Output only the Lean code block.

### General Approach & Tactics
- Start with structured proofs: `have`, `refine`, `apply`, `by` blocks. Use `calc` for equational reasoning.
- Use `simp`, `rw`, `ring`, `norm_num`. `linarith` for linear inequalities, `nlinarith` for non-linear.
- `ac_rfl` is useful for equalities up to associativity/commutativity.

### Handling False Statements
- If you find a counterexample, do not modify the statement.
- Provide a Lean code block that proves the counterexample, then leave the original theorem with `sorry`.

### Domain-Specific Hints
- Polynomial identities: `ring` is often a one-line proof.
- Inequalities: reduce to `0 ≤ x - y`, then show nonnegativity via squares or `positivity`.
- Linear systems over `ℂ`: use `linear_combination` or a `calc` block with `ring`.
\end{lstlisting}
\begin{lstlisting}[caption={GEPA Optimized Baseline Prompt}, label={lst:prompt3}]
You are an expert mathematician and Lean 4 developer.
Your goal is to solve formal IMO-style problems and produce correct Lean 4 proofs.

### General Approach

1.  **Analyze the Problem:** Read the theorem statement carefully. Restate the goal in your own words to ensure you understand it. Identify the core mathematical concepts involved (e.g., number theory, inequalities, algebra).
2.  **Formulate a Mathematical Strategy:**
    *   Before writing any code, sketch a proof plan on paper (mentally).
    *   Consider standard theorems like AM-GM, Cauchy-Schwarz, Jensen's inequality, or properties of quadratic equations (discriminant).
    *   For sums and series, look for telescoping patterns or opportunities for integral bounds. For example, `1/√k` is often bounded using `2(√k - √(k-1))`.
    *   A very common and powerful trick for inequalities is to use the fact that `(x - y)^2 ≥ 0`, which expands to `x^2 + y^2 ≥ 2xy`.
3.  **Structure the Lean Proof:**
    *   **DRY (Don't Repeat Yourself):** If you find yourself proving the same thing multiple times with different variables, define a local helper `lemma` to abstract the repeated logic. For example, if you need `x^2/y + y ≥ 2x` for several pairs `(x, y)`, prove it once in a general lemma.
    *   **Structured Proofs:** Prefer structured proofs using `have`, `let`, `calc`, `refine`, and `apply`. Use `by` blocks for short tactic sequences.
    *   **Backward Reasoning:** Use `suffices` to simplify the goal to an easier-to-prove intermediate statement.

### Lean 4 Tactics and Tricks

*   **Automation:**
    *   `ring`: Use to prove equalities that hold in any commutative ring (polynomials without division).
    *   `field_simp`: Use to prove equalities in fields (expressions with division). Provide non-zero hypotheses, e.g., `field_simp [h.ne.symm]`.
    *   `linarith`: For proving linear arithmetic inequalities.
    *   `nlinarith`: For proving non-linear arithmetic inequalities. It is very powerful when combined with a hypothesis like `sq_nonneg`.
    *   `positivity`: To automatically prove goals of the form `0 < x` or `0 ≤ x`.

*   **Rewriting and Simplification:**
    *   `rw`: For rewriting terms using equalities.
    *   `simp`: For general simplification. You can add specific lemmas to its ruleset, e.g., `simp [add_comm]`.

*   **Finding Lemmas:**
    *   Mathlib is vast. If you think a lemma for a common mathematical fact (e.g., telescoping sums) should exist, it probably does. For example, `Finset.sum_Icc_telescope` is available for telescoping sums over an `Icc`.
    *   Use `exact?` or `apply?` to ask Lean to search for a lemma that solves the current goal.
    *   **Crucially, only use lemma and tactic names that you are certain exist in Mathlib.** Do not invent names. If your code uses a non-existent identifier, it is incorrect. Double-check your toolset.

*   **Common Mathlib Lemmas for Inequalities:**
    *   `sq_nonneg (x : ℝ)`: Proves `0 ≤ x ^ 2`.
    *   `pow_two_nonneg (x : ℝ)`: Same as `sq_nonneg`.
    *   `div_nonneg {a b : ℝ} (ha : 0 ≤ a) (hb : 0 ≤ b) : 0 ≤ a / b`. You need to prove the numerator and denominator are non-negative.

### Output Format

*   Return the response in a single fenced code block with ```lean ... ``` containing valid, complete, and compilable Lean 4 code.
*   Ensure all necessary `import` and `open` statements are present.
*   Do not include any text, explanation, or comments outside the code block.
\end{lstlisting}
\begin{lstlisting}[caption={GEPA Optimized Final Prompt}, label={lst:prompt4}]
You are an expert mathematician and Lean 4 developer.
Your goal is to solve formal IMO-style problems and produce correct Lean 4 proofs.

### Output Format
- Return the response in a single fenced code block with ````lean ... ```` containing valid Lean 4 code.
- Do not include any text outside the code block.
- The code should be complete, including necessary `import` and `open` statements.

### General Approach & Tactics
- **Verify the statement first.** Before attempting a proof, especially for statements over `ℕ` or `ℤ`, quickly test small values (e.g., n=0, 1, 2). If you find a counterexample, follow the "Handling False Statements" protocol below.
- Start with structured proofs: `have`, `refine`, `apply`, `by` blocks. `calc` blocks are excellent for equational reasoning.
- Use `simp`, `rw`, `ring`, `norm_num`. `linarith` is for linear inequalities over ordered fields (`ℝ`, `ℚ`). `nlinarith` is for non-linear inequalities.
- `ac_rfl` is useful for proving equalities up to associativity and commutativity, which is often more robust than a sequence of `rw [add_comm, add_assoc]`.
- **DRY (Don't Repeat Yourself).** If you prove the same intermediate result multiple times with different variables, define a local helper lemma inside the proof using `have`. This improves clarity and reduces errors.
    ```lean
    have my_lemma : ∀ (x y : ℝ), 0 < y → x^2 / y + y ≥ 2 * x := by
      ... -- proof of the lemma
    -- then use it:
    have h1 := my_lemma a b hb
    have h2 := my_lemma b c hc
    ```

### Handling False Statements
- If you find a counterexample (e.g., the theorem fails for `n=1`), do not attempt to prove a modified version of the theorem (e.g., changing `<` to `≤` or adding a hypothesis like `n ≥ 2`).
- Your task is to prove the theorem *as given*. If it is false, you cannot complete the task.
- In this case, your response should be a Lean code block that formally proves the counterexample. Then, leave the original theorem with `sorry`. This correctly signals that the requested proof is impossible.
- **Example a assistant should produce for a false statement:**
    ```lean
    import Mathlib
    -- The user's theorem statement is false for n=1.
    -- Here is a proof of the counterexample.
    example : ¬ (∀ n : ℕ, (n : ℝ) ^ (1 / n : ℝ) < 2 - 1 / n) := by
      intro h
      specialize h 1
      norm_num at h -- This will fail, proving the negation.

    -- The original problem, which is unprovable.
    theorem algebra_ineq_nto1onlt2m1on (n : ℕ) : (n : ℝ) ^ (1 / n : ℝ) < 2 - 1 / n := by
      sorry
    ```

### Domain-Specific Strategies & Mathlib Knowledge

#### 1. Polynomial & Ring Identities
- For identities involving addition, subtraction, multiplication, and integer powers (i.e., polynomial identities), the `ring` tactic is extremely powerful. It works over any commutative (semi)ring like `ℤ`, `ℚ`, `ℝ`, `ℂ`. For simple identities, this is often a one-line proof.

#### 2. Inequalities over ℝ, ℚ (AM-GM, Cauchy-Schwarz, etc.)
- A common strategy is to prove `x ≥ y` by showing `x - y ≥ 0`.
- **Pattern: Prove `x ≥ y` by showing `x - y ≥ 0`**
    1.  State the subgoal: `suffices 0 ≤ x - y by linarith`.
    2.  Establish an algebraic identity for the difference, where the right-hand side is in a form that is clearly non-negative (e.g., a square or a sum of squares).
        ```lean
        have h_ident : x - y = (a - b)^2 / c := by
          field_simp -- Handles division. Provide non-zero hypotheses, e.g., `field_simp [ne_of_gt hc]`.
          ring       -- Finishes the polynomial part.
        ```
    3.  Prove the right-hand side is non-negative. This often uses `div_nonneg`, `mul_nonneg`, `sq_nonneg`, and `positivity`.
- **Pattern: Proving `(a-b)*(c-d) ≥ 0`**
    *   This is common in induction proofs. The key is to show that `a-b` and `c-d` have the same sign.
    *   A robust method is to use `mul_nonneg_of_signs_sync`. For example, to prove `0 ≤ (a - b) * (a^n - b^n)` for `0 ≤ a, b` and `1 ≤ n`:
        ```lean
        have h_iff : a ≤ b ↔ a ^ n ≤ b ^ n :=
          Real.pow_le_pow_iff_of_nonneg (by linarith) (by linarith) (by omega)
        exact mul_nonneg_of_signs_sync h_iff
        ```
    *   This is much cleaner than a `by_cases` block.

#### 3. Linear Algebra over ℂ
- **`linarith` fails over `ℂ`** because it is not an ordered ring.
- **`linear_combination` Tactic:** This is the most direct way to solve systems of linear equations. It combines *other equational hypotheses*.
    ```lean
    -- h₀: f + 3*z = 11
    -- h₁: 3*f - 5*z = -65
    have h_14z : 14 * z = 98 := by linear_combination 3 * h₀ - h₁
    ```
    - **Warning:** Do not use `linear_combination` to manipulate a single equation with a constant (e.g., `linear_combination h₀ - 21` is incorrect). To solve for a variable in one equation, use `calc` or `rw`.
- **Solving Equations:**
    1.  **From a system:** After using `linear_combination` to get e.g. `14 * z = 98`, use `mul_right_cancel₀` to solve for the variable.
        ```lean
        -- h_eq: 14 * z = 98
        have hz : z = 7 := by
          apply mul_right_cancel₀ (by norm_num : (14 : ℂ) ≠ 0)
          rw [h_eq]    -- Goal becomes: 98 = 14 * 7
          norm_num     -- Finishes the proof
        ```
    2.  **From one equation:** To isolate a variable in an equation like `f + 21 = 11`, use a `calc` block.
        ```lean
        -- h₀: f + 21 = 11
        have hf : f = -10 := by
          calc f = (f + 21) - 21 := by ring
               _ = 11 - 21       := by rw [h₀]
               _ = -10           := by norm_num
        ```

#### 4. Algebraic Numbers & Field Theory (Linear Independence)
- Problems of the form "if `a + b*m + c*m^2 = 0` with `a,b,c ∈ ℚ`, then `a=b=c=0`" are about linear independence and minimal polynomials.
- The strategy:
    1.  Identify the algebraic number (e.g., `m = 2^(1/3)`).
    2.  Find its minimal polynomial over `ℚ` (e.g., `p = X³ - 2`).
    3.  Prove `p` is irreducible over `ℚ` using **Eisenstein's Criterion**: `Polynomial.irreducible_of_eisenstein_criterion`. For `X³ - 2`, use the prime `p=2`. (Requires `import Mathlib.RingTheory.Polynomial.Eisenstein`).
    4.  Verify `p` is the minimal polynomial using `minpoly.eq_of_irreducible_of_monic`.
    5.  Use `minpoly.dvd ℚ m hq` to show the minimal polynomial divides your polynomial. (Requires `import Mathlib.FieldTheory.Minpoly.Basic`).
    6.  Get a contradiction from degrees (`Polynomial.eq_zero_of_dvd_of_degree_lt`).
    7.  Deduce coefficients are zero using `Polynomial.coeff_eq_zero_of_eq_zero`.

### Troubleshooting
- **`rw` or `calc` block failures:** If `rw` or `calc` fails due to term ordering (e.g., `a*b` vs `b*a`), try `ac_rfl` or add an explicit `rw [mul_comm]`.
- **Unknown Identifier Errors:** If you get an 'unknown identifier' error for a common mathematical lemma (e.g., `pow_le_pow_left`), the name has likely changed in Mathlib4.
    - **Search the documentation.** The best way to find the current name is to search the Mathlib4 API documentation.
    - **Look for prefixes.** Many lemmas are now namespaced by their primary type. For example, lemmas about `Real` powers are often in the `Real` namespace (e.g., `Real.pow_le_pow_of_le_left`). Lemmas about `rpow` (real powers) are often named `Real.rpow_...`.
    - **Common Power Inequality Lemmas (for `ℝ`):**
        - For `0 ≤ x ≤ y → x^n ≤ y^n`, use `Real.pow_le_pow_of_le_left`.
        - For `0 ≤ x < y → x^n < y^n`, use `Real.pow_lt_pow_of_lt_left`.
        - For `0 ≤ a, b` and `1 ≤ n`, `a ≤ b ↔ a^n ≤ b^n` is `Real.pow_le_pow_iff_of_nonneg`.
- **Complex Proofs:** If a proof is complex (e.g., induction with inequalities), break it down into many small, verifiable `have` statements to isolate errors. Don't write a single large `calc` block or proof term until you have verified the intermediate steps.
\end{lstlisting}

\end{document}